\documentclass[letterpaper, 10 pt, conference]{ieeeconf}  
\usepackage{amsmath, amssymb, cite}
\usepackage{makecell}
\IEEEoverridecommandlockouts                              
\usepackage{amsmath}
\usepackage{algorithm}
\usepackage[noend]{algpseudocode}
\usepackage{graphicx}
\usepackage{stfloats}
\usepackage{subfigure}
\usepackage{caption}
\usepackage{setspace}
\usepackage[colorlinks,linkcolor=blue]{hyperref}

\usepackage{lipsum}

\newcommand\blfootnote[1]{%
\begingroup 
\renewcommand\thefootnote{}\footnote{#1}%
\addtocounter{footnote}{-1}%
\endgroup 
}

\title{\LARGE \bf
OminiAdapt: Learning Cross-Task Invariance for Robust and Environment-Aware Robotic Manipulation
}

\author{Yongxu Wang$^{1*}$, Weiyun Yi$^{2*}$, Xinhao Kong$^{3*}$, Wanting Li$^{\dag}$} 

\begin{document}


\twocolumn[{%
\renewcommand\twocolumn[1][]{#1}%
\maketitle

\begin{figure}[H]
    \makebox[\textwidth][c]{%
        \includegraphics[width=\textwidth]{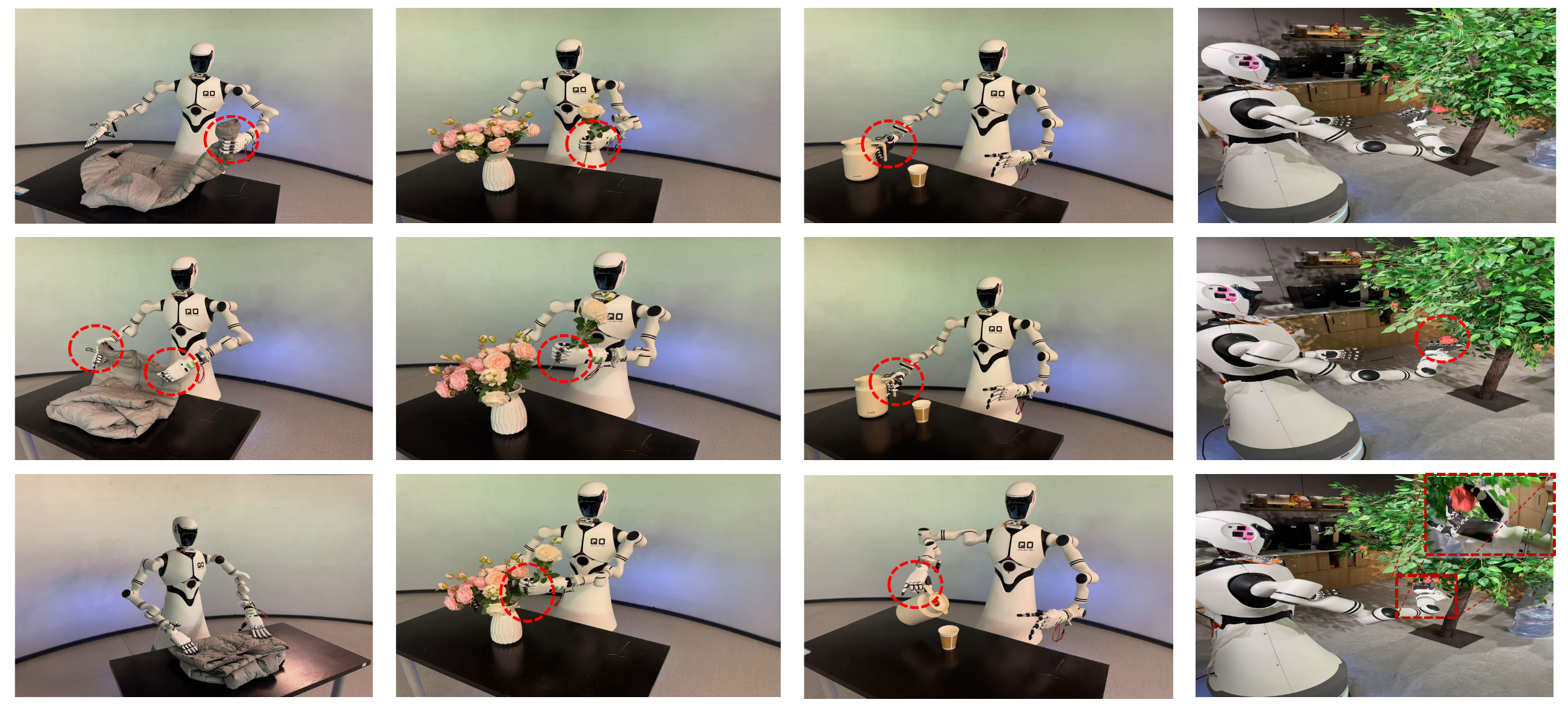}
    }
    \makebox[\textwidth][c]{%
        \parbox{\textwidth}{\centering\caption{Results of our method in different tasks.}}
    }
    \label{fig:Alltask}
\end{figure}
}]

\thispagestyle{empty}
\pagestyle{empty}

\blfootnote{$^{1}$University of Science and Technology of China, Hefei, China.}
\blfootnote{$^{2}$Beijing Institute of Technology, Beijing, China.}
\blfootnote{$^{3}$Beijing Institute of Technology, Beijing, China.}
\blfootnote{$^{\dag}$Institute of Automation Chinese Academy of Sciences, Beijing, China.}
\blfootnote{$^{*}$Work was done during the internship at the Institute of Automation Chinese Academy of Sciences.}


\begin{abstract}

With the rapid development of embodied intelligence, leveraging large-scale human data for high-level imitation learning on humanoid robots has become a focal point of interest in both academia and industry. However, applying humanoid robots to precision operation domains remains challenging due to the complexities they face in perception and control processes, the long-standing physical differences in morphology and actuation mechanisms between humanoid robots and humans, and the lack of task-relevant features obtained from egocentric vision. To address the issue of covariate shift in imitation learning, this paper proposes an imitation learning algorithm tailored for humanoid robots. By focusing on the primary task objectives, filtering out background information, and incorporating channel feature fusion with spatial attention mechanisms, the proposed algorithm suppresses environmental disturbances and utilizes a dynamic weight update strategy to significantly improve the success rate of humanoid robots in accomplishing target tasks. Experimental results demonstrate that the proposed method exhibits robustness and scalability across various typical task scenarios, providing new ideas and approaches for autonomous learning and control in humanoid robots. The project will be open-sourced on GitHub.

\end{abstract}

\section{INTRODUCTION}



In recent years, the development of humanoid robots and embodied intelligence has progressed rapidly. However, a significant challenge remains in enabling these robots to autonomously make decisions and complete tasks in unstructured environments\cite{lu2024Manigaussian}. Among these complexities, dexterous manipulation of the robot's upper body stands out as a critical aspect, as it is essential for humanoid robots to understand their surroundings, make autonomous decisions, and perform tasks effectively using their mechanical systems and controls.

Recently, several approaches have been proposed to address the challenge of robotic decision-making for manipulation tasks. These include reinforcement learning on real robots, reinforcement learning in simulation, and imitation learning on real robots. Among these, real-robot imitation learning has gained considerable attention due to its efficiency in learning specific manipulation tasks, as it eliminates the need to bridge the gap between simulation and real-world deployment. For example, Mobile ALOHA \cite{fu2024mobile} introduced the Action Chunking Transformer (ACT) algorithm, which divides action sequences into fixed-length blocks (e.g., 50-step action blocks) and uses a Transformer model to predict these action blocks. This method reduces the complexity of sequential modeling while preserving trajectory coherence. Additionally, HumanPlus \cite{fu2024humanplus} developed a decoder-only policy called HIT, which facilitates low-cost learning of manipulation tasks with minimal human demonstration data, by leveraging a hierarchical control architecture on a real humanoid robot.



Despite these advancements, critical bottlenecks hinder the practical deployment of imitation learning in humanoid robots. For instance, egocentric vision, which serves as a primary sensory input for imitation, is susceptible to dynamic background clutter and variations in target objects, leading to inconsistent feature extraction and shifts in perceptual domains. Second, physical discrepancies between training simulations and real-world environments introduce action execution errors, as policies trained on idealized models fail to account for real-world dynamics. Third, traditional normalization techniques like batch normalization, optimized for fixed training distributions, struggle to adapt to novel conditions in real-time, leading to performance degradation during online operation. These intertwined challenges in perception robustness, physical consistency, and dynamic adaptation collectively limit the scalability and reliability of imitation learning frameworks for humanoid robots.

To address these issues, we propose a multimodal adaptive imitation learning framework named OminiAdapt for humanoid robots, with three key contributions:

1) \textbf{Multi-dimensional Attention Enhancement}: Based on the Convolutional Block Attention Module, the algorithm dynamically allocates spatial and channel weights to visual feature maps, enhancing the robust representation of hand-object interaction features while suppressing irrelevant background noise;

2) \textbf{Dynamic Hand-Object Segmentation}: Through real-time tracking and hand-object segmentation, our method generates pixel-wise masks to filter and eliminate dynamic background distractions, improving decision-making focus;

3) \textbf{Dynamic Adaptive Batch Normalization (DABN)}: Our method adjusts normalization parameters using online inference data, effectively mitigating feature distribution shifts across tasks.


\section{RELATED WORKS}

\subsection{Imitation Learning in robotics manipulation}
Imitation learning enables robots to learn skills from expert demonstrations\cite{NIPS1988_humanlearn}, with Behavioral Cloning (BC) being a foundational approach. BC directly maps observed states to actions but struggles with action drift, where small errors compound over time. Recent advancements seek to address this limitation through more complex models.

Fu et al.\cite{fu2024mobile} introduce the Action Chunking Transformer (ACT), which decomposes tasks into interpretable action chunks using a Transformer-based architecture, improving long-horizon task performance by mitigating drift. However, like BC, ACT is still sensitive to environmental variations, requiring retraining when faced with new conditions. Subsequent work by HumanPlus\cite{fu2024humanplus} introduced the HIT algorithm, which uses a decoder-only Transformer to predict action sequences autoregressively. HIT improves unstructured environments, such as household tasks, by employing sparse attention to focus on relevant input features. Despite its robustness, HIT’s performance still degrades with significant changes in the environment, as it relies heavily on the training data’s coverage, requiring additional demonstrations when facing new scenarios.

Beyond Transformer-based approaches, Diffusion Models have also been explored for imitation learning, offering a generative approach to action prediction. These models generate multimodal action distributions through iterative denoising, showing promise in noisy data environments. The Diffusion Policy \cite{chi2024diffusionpolicy} pioneered the application of diffusion networks to real-world trajectory prediction, employing stochastic differential equations to model action sequence latent spaces. The OCTO \cite{octo_2023} framework further integrates attention mechanisms into diffusion models, enabling prioritized processing of critical action nodes. However, their computational overhead (e.g., numerous denoising steps) introduces significant latency and limits real-time applicability.

Despite these advancements, environmental adaptability remains a fundamental limitation. Current methods often require retraining with new demonstrations to handle minor environmental variations, underlining the need for more robust domain adaptation techniques to improve task transferability and reduce data dependence.

\subsection{Application of Attention Mechanisms in Dexterous Robotic Manipulation}  
In recent years, the success of multi-semantic spatial structures in computer vision has provided new insights into robotic manipulation tasks. In complex manipulation scenarios, robots must simultaneously understand both the global structure of the environment and local details. Effectively integrating these two types of information can significantly enhance the precision of target capture and manipulation. For example, the InceptionNets series \cite{szegedy2017inception,szegedy2015going,szegedy2016rethinking} pioneered the use of parallel multi-scale convolution branches to capture features at different receptive fields, thereby laying the groundwork for subsequent methods that leverage spatial priors to strengthen feature representation. Similarly, SKNet \cite{li2019selective} further combined multi-scale convolutions with channel attention by introducing the squeeze-and-excitation mechanism from SENet \cite{hu2018squeeze}, effectively integrating global and local spatial information. This approach facilitates the robust recognition of critical manipulation regions in complex robotic environments.



Meanwhile, to alleviate the memory and computational burdens introduced by high-dimensional attention, various attention decomposition methods have emerged. For instance, CA \cite{hou2021coordinate} and ELA \cite{xu2024ela} perform unidirectional spatial compression along the height (H) and width (W) axes, respectively, partially preserving spatial structures. Similarly, SA \cite{zhang2021sa} and EMA \cite{ouyang2023efficient} partition high-dimensional features into subcomponents to reduce computational complexity. Additionally, CPCA \cite{huang2024channel} applies stripe convolutions to decrease the number of parameters in large-kernel convolutions, while recent work such as RMT \cite{fan2024rmt} decomposes MHSA along the H and W dimensions to control the exponential growth in computational cost. Together, these strategies not only improve the efficiency of attention modules in practical applications but also establish an optimized synergy between global and local information. This synergy effectively mitigates the distribution shift between expert demonstrations and practical operation strategies in imitation learning—an approach analogous to the methods employed by Mamba \cite{gu2023mamba} and VMamba \cite{liu2025vmamba}, which capture global contextual information through scanning mechanisms and offer new breakthroughs in enhancing both the real-time performance and accuracy of robotic manipulation systems.

\section{METHOD}
\newcommand{\concat}{\oplus}  
\newcommand{\TD}{\mathrm{TransformerDecoder}}
\newcommand{\MLP}{\mathrm{MLP}}

\begin{figure*}

\vspace{-0.2in}
    \centering
    \includegraphics[width=1\linewidth]{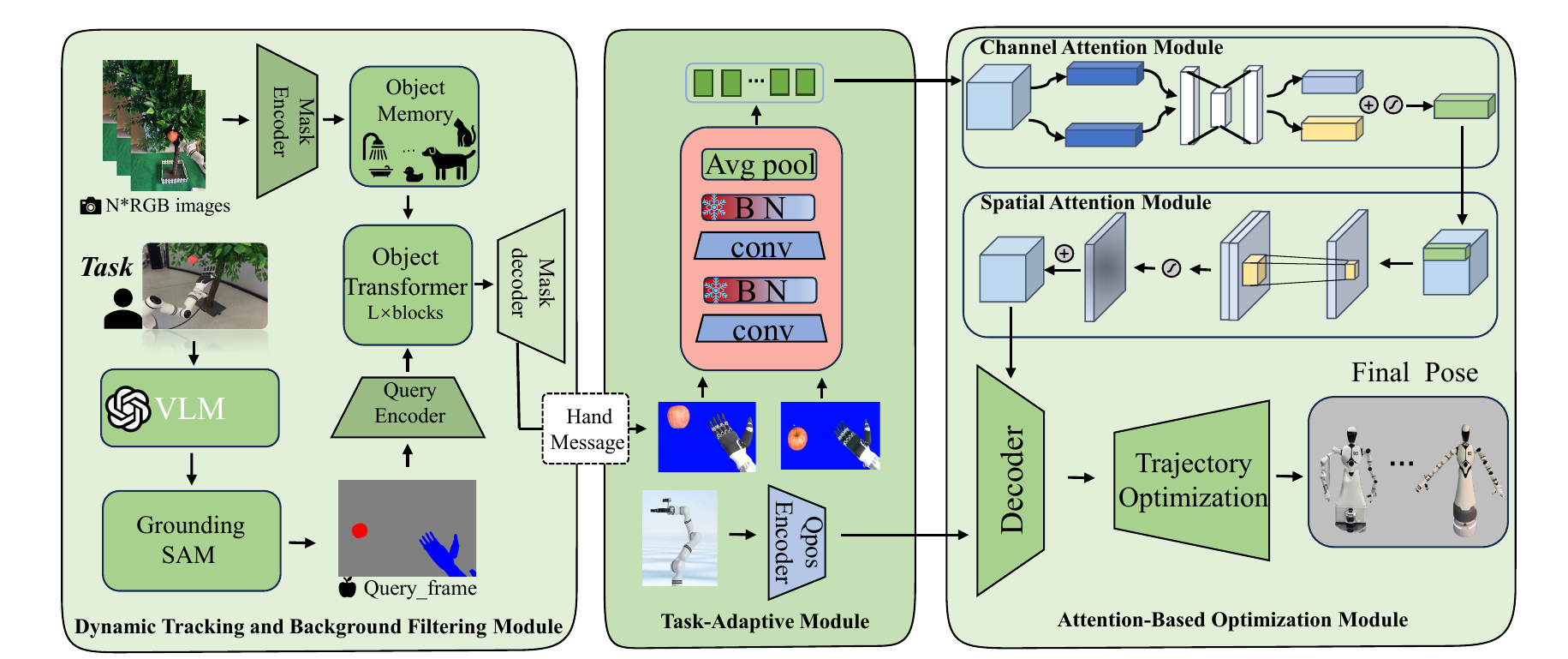}
    \vspace{-0.2in}
    \caption{OminiAdapt Overview. The first frames from $N$ viewpoints are processed by the task interpreter module based on VLM and GorungdingSAM to distinctly generate an initial query frame for every view that initializes the tracking algorithm. Subsequently, the RGB video streams from all viewpoints undergo continuous object tracking, where key elements are masked to filter out background information irrelevant to the task. Then, the image features are extracted using a semi-frozen backbone with dynamic adaptive batch normalization (BN) layers, and features are enhanced through a channel-space attention module. The enhanced features, along with the embedded robot's proprioceptive states are fed into the decoder of a Transformer architecture to predict the robot's actions over the next $T$ steps, with trajectory smoothing applied for improved motion consistency.}
    \vspace{-0.2in}
    \label{fig:overview}
\end{figure*}

\subsection{Overview}
In order to cope with the high sensitivity of imitation learning to the environment and the huge cost of retraining for new tasks, we propose OminiAdapt, an imitation learning-based robotic policy architecture, which is illustrated in Fig.~\ref{fig:overview}. 
Our system employs $N$ cameras (default $N=3$ at the head, chest, wrist, etc.) to capture synchronized image streams $\{ I_t^{v_i} \}_{i=1}^N \in \mathbb{R}^{N \times3 \times {H} \times {W}}$ at $30\,\text{Hz}$. Valid frame acquisition is triggered when the robotic arm's motion energy exceeds $\lVert \Delta q^{\text{arm}}_t \rVert_2^2 + \lVert \Delta\theta^{\text{hand}}_t \rVert_2^2 > 0.1$, ensuring sample efficiency. Given different RGB camera views and robot proprioception $s_t = [q_{t}^{\text{arm}}, \theta^{\text{hand}}_t] \in \mathbb{R}^{13}$ at current time $t$, OminiAdapt predicts next T-steps of robotic arm and hand trajectory. Here are detailed descriptions of the process.

Firstly, we use the task interpreter module and zero-shot semantic segmentation module to initialize the image preprocessing, and subsequent images $\{ I_t^{v_i}\}$ use masked images of the previous moment as query images to generate masks $\{M_{t}^{v_i}\}$ by continuous object tracking.
Secondly, the mask images are processed by a ResNet18 backbone with frozen convolutional weights and trainable dynamic batch normalization (BN) layers, extracting view-specific characteristics $\{F_t^{v_i}\}_{i=1}^N\in\mathbb{R}^{N \times C \times H \times W}$, making it faster to migrate learning to a new task.
Thirdly, these feature maps are enhanced through a channel-space attention module named CBAM \cite{woo2018cbam}, which contributes to policy robustness when facing different environmental disturbances without large amounts of data needing to be collected even if there is only a small change in the environment. The enhanced multi-view features are aggregated through:
\begin{equation}
\mathcal{F}_{t}^{\text{3D}} = \text{Stack}\left( \{\text{CBAM}(F_{t}^{v_i} )\}_{i=1}^N \right) 
\end{equation}
where $\text{CBAM}(\cdot)$ denotes the attention operation. 
Finally, the fused feature $\mathcal{F}^{\text{in}}_{t} = \text{Concat}(\mathcal{F}_{t}^{\text{3D}}, \mathcal{F}_{t}^{\text{pro}})$ is fed into a transformer decoder to predict $T$-step future actions $\{\tau_{t+k}^{\text{pred}}\}_{k=1}^T$, where $\mathcal{F}_{t}^{pro}=MLP(s_t)$ is embedded robot proprioception at time $t$. Following \cite{fu2024mobile}, we adopt chunk-wise prediction for temporal coherence. The transformer decoder outputs actions:
\begin{equation}
\{\tau_{t+k}^{\text{pred}}\}_{k=1}^T = Transformer\left( \mathcal{F}_{t}^{\text{3D}} \concat \MLP(s_t) \right)
\end{equation}
where
$\tau_{t+k}^{\text{pred}} = [q_{t+k}^{\text{arm,pred}}, \theta_{t+k}^{\text{hand,pred}}]$ ,
optimized by the loss:
\begin{equation}
\mathcal{L} = \sum_{k=1}^T 0.95^{T-k} \left( \lVert \tau_{t+k}^{\text{pred}} - \tau_{t+k}^{\text{gt}} \rVert_2 + 0.2 \lVert \nabla \tau_{t+k}^{\text{pred}} \rVert_2 \right)
\end{equation}
The first term ensures trajectory accuracy, while the second term penalizes velocity discontinuities. The exponential decay factor $0.95^{T-k}$ implements curriculum learning.

\subsection{Continuous Object Tracking for Visual Preprocessing}
In preprocessing, the first frame image in all views initializes with zero-shot semantic segmentation using a task interpreter module and Grounding-SAM \cite{ren2024grounded}(ViT-B backbone) for object localization, and the subsequent frames are processed by continuous object tracking and masking through Cutie tracker\cite{cheng2023putting}.
\begin{figure}
    \centering
    \includegraphics[width=1\linewidth]{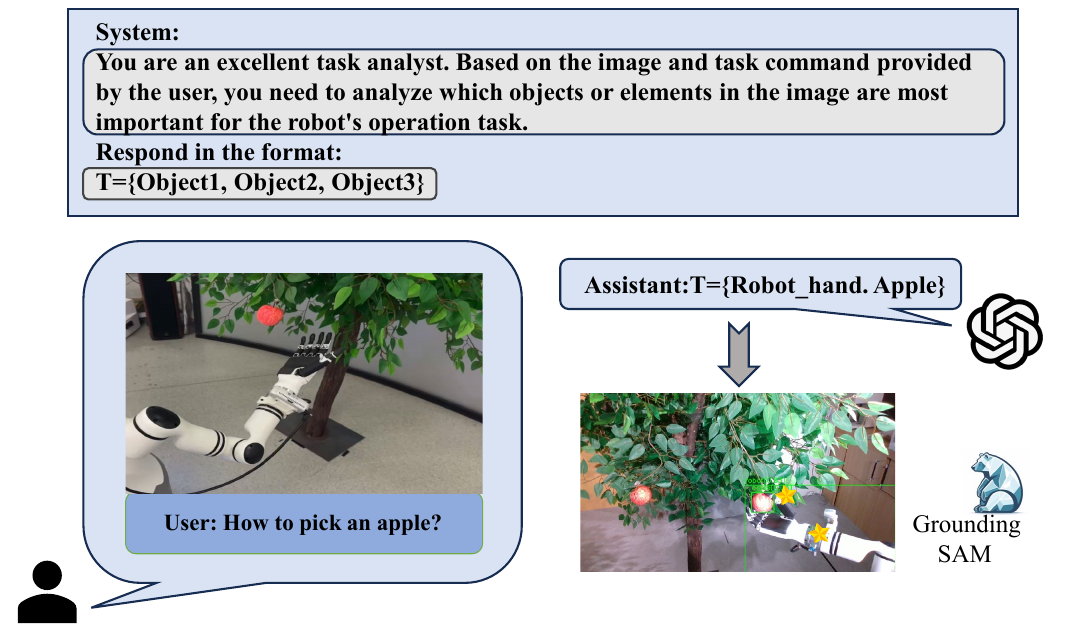}
    \vspace{-0.2in}
    \caption{Task Interpreter Module}
    \vspace{-0.2in}
    \label{fig:vlm}
\end{figure}
As shown in Fig.~\ref{fig:vlm}, the input of a multimodal large model consists of camera-captured images from different task scenarios and the corresponding task descriptions. The model identifies the objects or elements in the image that the robot needs to focus on most from different perspectives and generates text-based prompts. Based on this operation, we present a generation paradigm for automatic key element prompts.
Given these generated text prompts, GroundingSAM is used to perform zero-shot semantic segmentation on the initial frame images ${I}_0^v$ from each perspective, generating the corresponding masks \( M_0^v = \text{GroundingSAM}(I_0^v, T) \) to initialize the image preprocessing. The current image is then used as the query frame for the subsequent frame, and a continuous loop of object tracking, segmentation, and mask processing is performed.
Subsequent frames are processed through an enhanced CUTIE tracker to perform continuous object-level tracking and segmentation. This process not only filters out background interference but also morphological dilation ($r=15$ pixels) expands hand masks for occlusion robustness, trying to preserve edge information of the robotic arm and hand, even if occluded by objects like leaves or clothing. To ensure consistency across different viewpoints, we apply an inter-view consistency constraint, which forces high alignment of masks across different views after geometric transformations. The constraint is expressed as:
\begin{equation}
\frac{1}{N}\sum_{v_i \neq v_j} \text{IoU}\left( \mathcal{M}_t^{v_i}, \mathcal{W}(\mathcal{M}_t^{v_j}) \right) > 0.65
\label{eq:iou}
\end{equation}
where \( N\) is the number of viewpoints and $\mathcal{W}(\cdot) $ represents the coordinate transformation matrix between views. 
\subsection{Dynamic Adaptive Batch Normalization}
The feature extraction architecture consists of a pre-trained feature extraction module and a channel-space attention module to enhance focus on hand-object interaction. The backbone network, employing a dynamic adaptive batch normalization (DABN) design, is a frozen multi-view shared-weight ResNet18 (pre-trained on ImageNet), where only the BN layers in the backbone are activated to be jointly fine-tuned during the task-specific training. This DABN design allows the backbone to reduce the gap between different tasks while sharing convolution layer parameters and adjusting the BN parameters to easily start and quickly adapt to new tasks, reducing training time and resource consumption, and improving policy transferability and environmental adaptability. Each task strategy maintains its own set of independent BN parameters \( \{ \mu_k, \sigma_k, \gamma_k, \beta_k \} \). The forward propagation is as follows:
\[
\hat{x} = \frac{x - \mu_k}{\sigma_k^2 + \epsilon} \cdot \gamma_k + \beta_k
\]
After overall training, task-specific embedding vectors \( e_{\text{task}} \in \mathbb{R}^{64} \) are routed to select the BN parameters:
\[
[\gamma_k; \beta_k] = \text{MLP}(e_{\text{task}})
\]
\subsection{Multi-dimensional Attention Mechanism}
The extracted features from different viewpoints are passed through the channel-space attention module, we employed the CBAM Module in \cite{woo2018cbam}, which suppresses task-irrelevant interference, high-frequency noise, and irrelevant color channels and contributes to focusing on spatial relationships between the hand and object, particularly in the interaction region, optimizing object contours and hand parts.
The channel attention mechanism focuses on suppressing irrelevant feature channels, improving the model's resistance to high-frequency noise (e.g., robot reflections), which computes weights $\alpha_c \in \mathbb{R}^C$ via dual-path pooling (GAP and GMP).
The spatial attention mechanism generates heatmaps through convolutional layers, which allows the policy to better learn the spatial relationship between the hand and task objects, focusing on details in hand-object interaction, which efficiently improves its ability to handle background interference and enhances robustness and environmental adaptability. 

\section{EXPERMENT}
\subsection{Hardware and Experimental setting}
Our dual-arm mobile robot consists of an AGV chassis and a pair of 7-DOF arms equipped with different dexterous hands. The robotic arms are RM75-6F lightweight humanoid arms from Realman, integrated with six-axis force sensors. The right arm features an RH56DFX dexterous hand from Inspire, while the left arm is equipped with an L10 dexterous hand from LinkerHand, which possesses ten active degrees of freedom. Both hands are controlled via serial communication. The AGV chassis is the WATER2 model from YUNJI. All components are connected to an NVIDIA Jetson Orin through Ethernet.The robot's head, neck, and right hand are equipped with three RGB cameras (Intel RealSense D435i). The Inspire dexterous hand can exert a maximum force of 10 N, while the LinkerHand L10 offers enhanced manipulation capabilities through its multi-DOF design and features tactile sensors on all five fingertips. The arms have a rated payload of 5 kg. An illustration of our robot is shown in the figure below. In future work, we plan to fully leverage the tactile data from the LinkerHand to further improve our approach.

\begin{table}[ht]
\centering
\begin{tabular}{@{}ccccc@{}}
\hline
\textbf{Methods} & \makecell{\textbf{Fold} \\ \textbf{clothes}} & \makecell{\textbf{Pick} \\ \textbf{apples}}  & \makecell{\textbf{Arrange} \\ \textbf{flowers}} & \makecell{\textbf{Pour} \\ \textbf{water}} \\ \hline
Demos & 40 & 50 & 60 & 60 \\
\hline
\end{tabular}
\vspace{0.1in}
\caption{The amount of training data set for different scenarios.}
\vspace{-0.1in}
\label{tab:amount}
\end{table}

In the comparative experiments, we collected 40 data samples for each task in the folding clothes scenario and 50 training samples in the apple-picking scenario. For the more challenging tasks, such as pouring water and arranging flowers, we collected 60 data samples,as shown in Table .\ref{tab:amount} \textbf{Each scenario was evaluated through 20 experimental runs to compare success rates. In the ablation experiments, each task was attempted 10 times}, with both step-by-step success rates and final success rates analyzed for each trial. Additionally, for multi-scenario generalization tasks, experiments were conducted across different scenarios using varying data distribution ratios for testing.
\begin{figure*}

\vspace{-0.2in}
    \centering
    \includegraphics[width=1\linewidth]{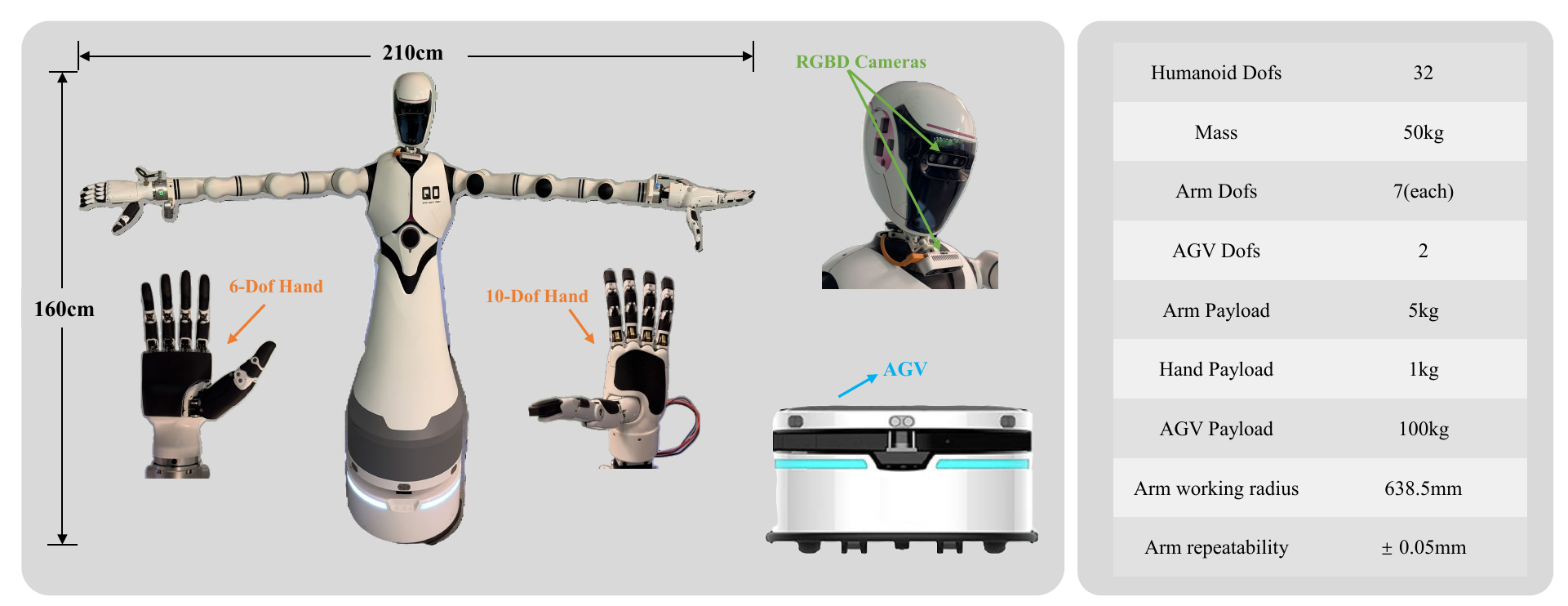}
    \vspace{-0.1in}
    \caption{This figure presents our robot's hardware configuration. The upper right shows the complete system featuring dual Realman RM75-6F robotic arms equipped with an Inspire RH56DFX dexterous hand and a LinkerHand L10 dexterous hand. The perception system comprises three Intel RealSense D435i cameras mounted on a Yunji Water2 mobile platform.}
    \vspace{-0.2in}
    \label{fig:hardware}
\end{figure*}

\subsection{Comparison Experiments}

As shown in Table .\ref{tab:t1}, we compared our model with the HIT\cite{fu2024humanplus} and ACT\cite{fu2024mobile} algorithms across various tasks, including folding clothes, picking apples, arranging flowers, and pouring water. Our model demonstrated superior performance in most tasks, particularly in apple picking and flower arranging. For instance, in the apple picking task, our model achieved a remarkable success rate of 90\%, significantly outperforming HIT and ACT, which both achieved 35\%. This improvement can be attributed to our model's advanced grasping strategies and enhanced object detection capabilities. Similarly, in the flower arranging task, our model achieved an 65\% success rate, compared to HIT's 25\% and ACT's 20\%, showcasing the effectiveness of our precise motion control and high-resolution visual processing.


\begin{table}[ht]
\centering
\begin{tabular}{@{}ccccc@{}}
\hline
\textbf{Methods} & \makecell{\textbf{Fold} \\ \textbf{clothes}} & \makecell{\textbf{Pick} \\ \textbf{apples}}  & \makecell{\textbf{Arrange} \\ \textbf{flowers}} & \makecell{\textbf{Pour} \\ \textbf{water}} \\ \hline
HIT  & 90 & 35 & 25 & 20 \\
ACT  & 80 & 35 & 20 & 10 \\
Ours & \textbf{100} & \textbf{90} & \textbf{65} & \textbf{60} \\ \hline
\end{tabular}
\vspace{0.1in}
\caption{Comparison of experimental results between HIT and ACT in fine-grained tasks.}
\vspace{-0.1in}
\label{tab:t1}
\end{table}
In the task of folding clothes, our model achieved a perfect success rate of 100\%, surpassing HIT (90\%) and ACT (80\%). This highlights our model's ability to excel in repetitive, structured tasks. Furthermore, in the more delicate task of pouring water, our model achieved a 60\% success rate, significantly higher than HIT (20\%) and ACT (10\%). This demonstrates the robustness of our model in handling tasks requiring fine motor skills and precise control. Overall, the results indicate that our model  not only excels in intricate and delicate operations but also maintains strong performance in routine and repetitive tasks.

By conducting 20 trials in each task, we assessed the consistency and reliability of each model. Similarly, we calculated the success rate of each subtask by dividing the number of successful attempts by the total number of attempts. For example, in the apple picking task, if the robot successfully located the apple before attempting to pick it, that subtask was considered successful. Our model demonstrated its strong capabilities in tasks requiring high perceptual acuity and precision in movement, and it was able to compete with the top-performing algorithm HIT in more conventional and repetitive tasks.

\subsection{Ablation experiment}

In our research on imitation learning, we conducted detailed ablation experiments to assess strategies for enhancing model generalization and to evaluate the impact of various components on the task of tracking apples. The specific experimental designs included partial freezing of Batch Normalization (BN) layers, domain transfer tasks, and the application of the Cutie technique to the apple tracking task, which involved masking either the background or the apple. Additionally, we integrated the CBAM module, which includes spatial and channel attention, to enhance the model’s ability to recognize key features.

\textbf{1. Ablation Experiment Using Different Masks with Cutie}

\begin{figure}[ht]
\vspace{-0.1in}
    \centering
    \includegraphics[width=1\linewidth]{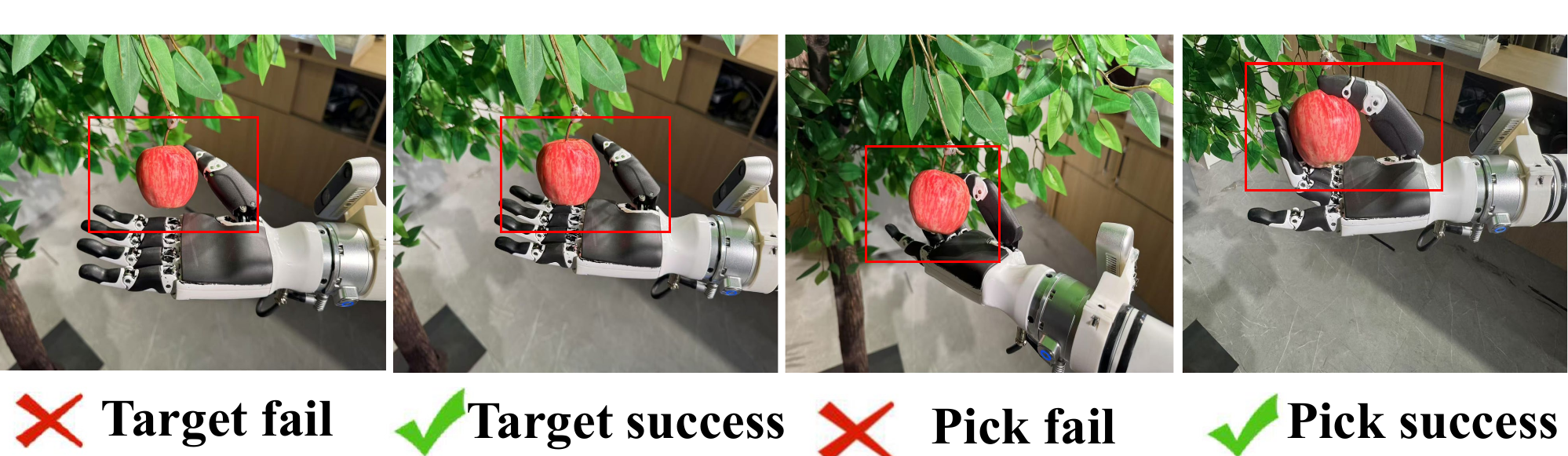}
    \vspace{-0.1in}
    \caption{Explanation of indicators for determining whether a task is successful or not.}
    \vspace{-0.1in}
    \label{fig:success}
\end{figure}

The Target attribute represents the accuracy of positioning. Successfully grasping the apple requires precise alignment with the middle position of the index finger. The Pick attribute indicates the success of the grasping action. Even with accurate positioning, factors such as the swaying of the apple may affect the success of the grasp. The grasping success rate reflects the overall success of the entire process. The evaluation metrics are shown in Figure \ref{fig:success}.

\begin{table}[ht!]
\centering

\begin{tabular}{@{}ccccccc@{}}
\hline
\textbf{Methods}   & \textbf{Target}& \textbf{Pick}  & \textbf{Success}& \textbf{OPSR} \\ \hline
Mask apple        & 70            &  70  & 60& 60                     \\
Mask hand        & 90  &  70   & 70          &60                 \\
Mask background   &\textbf{ 90}   &\textbf{80} & \textbf{80}  &\textbf{70}
 
\\ \hline

\end{tabular}
\vspace{0.1in}
\caption{Ablation Experiment Using Different Masks with Cutie}
\vspace{-0.1in}
\label{tab:t2}
\end{table}
To evaluate the impact of different masking strategies on model performance, we conducted an ablation study focusing on the Obstructive Picking Success Rate (OPSR) and other task metrics, as shown in Table. \ref{tab:t2}. The results reveal that applying a mask to the background yielded the best performance, achieving an OPSR of 70\% and an overall success rate of 80\%. This strategy effectively enhanced the model's ability to target the apple and execute the picking motion, with a significant improvement in the "Picking off" metric (80\%) compared to masking the apple or the hand. Notably, masking the apple directly resulted in the lowest success rate (60\%) and OPSR (60\%), suggesting that occluding the object of interest hinders recognition and reduces task performance. These findings highlight the importance of providing clear and relevant visual cues, as background masking allows the model to focus on task-critical elements, thereby improving both recognition and manipulation accuracy.

\textbf{2. Ablation Study on Different Attention Mechanisms}

\begin{figure}[ht]
\vspace{-0.1in}
    \centering
    \includegraphics[width=1\linewidth]{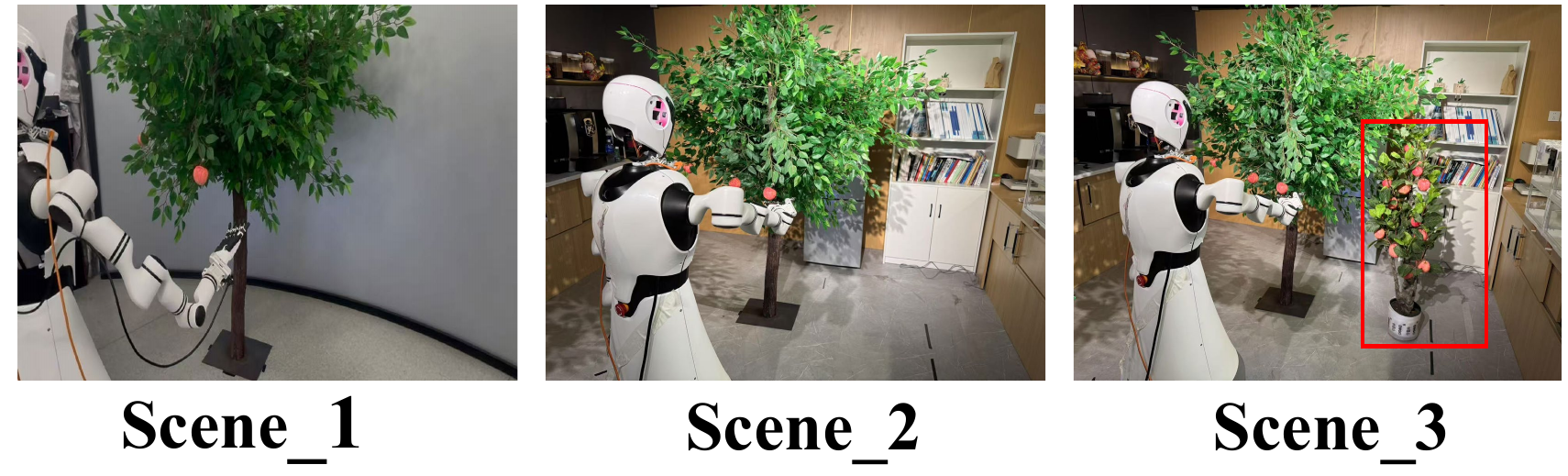}
    \vspace{-0.1in}
    \caption{Three scenarios of data collection and experiments}
    \vspace{-0.1in}
    \label{fig:scene}
\end{figure}

In the ablation study, we analyzed the impact of different attention mechanisms, including CAM, SAM, and CBAM, on the model's performance across various tasks,  As shown in Table.\ref{tab:t8}. \textit{We collected 25 data samples each in Scene 1 and Scene 2, and tested the generalization of the experiment to new environments in Scene 3}, as shown in Figure \ref{fig:scene}.The results demonstrate that removing any of these components significantly reduces the model's success rate and adaptability. For instance, without CBAM, the model's success rate dropped to 40\%, compared to 90\% with all components included. Additionally, the "New Scene" metric, which evaluates the model's adaptability in unfamiliar environments, showed a substantial decrease from 70\% with the full model to 40\% without CAM and 30\% without SAM. These findings emphasize the critical role of attention mechanisms in improving the model's ability to generalize to new scenarios and maintain high performance. Notably, our full model outperformed all ablated versions, achieving a success rate of 90\% and demonstrating robust performance across metrics like "Obstructive Grasp" (70\%) and "Dynamic Impact" (90\%). This underscores the importance of integrating advanced attention modules to enhance both task precision and adaptability.


 
 
\begin{table}[ht!]
\centering
\begin{tabular}{@{}cccc@{}}
\hline
\textbf{Methods} & \makecell{\textbf{Targeting} \\ \textbf{apple}} & \makecell{\textbf{Picking} \\ \textbf{off}} & \makecell{\textbf{Success} \\ \textbf{rate}} \\ \hline
Without CAM   & 70 & 60 & 60 \\
Without SAM   & 60 & 40 & 40 \\
Without CBAM  & 40 & 40 & 40 \\
Ours          & \textbf{100} & \textbf{90} & \textbf{90} \\
\hline
\textbf{Methods} & \makecell{\textbf{Obstructive} \\ \textbf{grasp}} & \makecell{\textbf{New} \\ \textbf{Scene}} & \makecell{\textbf{Dynamic} \\ \textbf{impact}} \\ \hline
Without CAM   & 40 & 40 & 30 \\
Without SAM   & 40 & 30 & 40 \\
Without CBAM  & 10 & 10 & 20 \\
Ours          & \textbf{70} & \textbf{70} & \textbf{90} \\
\hline
\end{tabular}
\vspace{0.1in}
\caption{ Ablation Study on Different Attention Mechanisms}
\vspace{-0.2in}
\label{tab:t8}
\end{table}

\textbf{3.Ablation Experiment on Partial BN Layer Freezing}

In this ablation study, we investigated the impact of utilizing Dynamic Adaptive Batch Normalization (DABN) layers on the model's performance,  As shown in Table. \ref{tab:t9}. DABN aims to enhance the model's adaptability by maintaining statistical consistency across domains. The results indicate that our model with DABN achieved superior performance in key metrics, particularly in the "Dynamic Impact" task, where the success rate improved from 70\% (without DABN) to 90\%. Additionally, in the "Targeting Apple" task, the inclusion of DABN allowed the model to achieve a perfect success rate (100\%), compared to 90\% without DABN. These improvements highlight that DABN effectively mitigates overfitting in the source domain while enhancing the model's generalization capabilities in dynamic and challenging environments. This demonstrates the critical role of adaptive normalization techniques in boosting both task-specific accuracy and domain robustness.





\begin{table}[ht!]
\centering
\begin{tabular}{@{}cccc@{}}
\hline
\textbf{Methods} & \makecell{\textbf{Targeting} \\ \textbf{apple}} & \makecell{\textbf{Picking} \\ \textbf{off}} & \makecell{\textbf{Success} \\ \textbf{rate}} \\ \hline
Without DABN   & 90  & 90  & 90 \\
Ours           & 100 & 90  & 90 \\ \hline
\textbf{Methods} & \makecell{\textbf{Obstructive} \\ \textbf{grasp}} & \makecell{\textbf{New} \\ \textbf{Scene}} & \makecell{\textbf{Dynamic} \\ \textbf{impact}} \\ \hline
Without DABN    & 70  & 70  & 70 \\
Ours            & 70  & 70  & 90 \\ \hline
\end{tabular}
\vspace{0.1in}
\caption{Ablation Experiment on Partial BN Layer Freezing}
\vspace{-0.2in}
\label{tab:t9}
\end{table}

\section{CONCLUSIONS}
This paper presents a multimodal adaptive imitation learning framework addressing perceptual domain shift, physical domain gap, and online adaptation challenges for high-DOF robots in unstructured environments. Our approach enhances model generalization through three key innovations: background-aware dynamic masking, multi-dimensional attention enhancement, and dynamic batch normalization.
Experiments demonstrate significant performance improvements in precision-demanding tasks including apple picking, garment folding, flower arrangement, and water pouring through our methods. These findings validate that our framework effectively addresses core challenges in robotic imitation learning, providing a reliable solution for complex task execution in dynamic, unstructured environments. Despite these advances, several limitations remain such as inconsistency of multiple perspectives and lack of tactile modalities. By addressing these limitations, our multimodal adaptive framework will further enhance robot performance across diverse environments, establishing a foundation for real-world intelligent robotic applications.

\addtolength{\textheight}{-12cm}   








\bibliographystyle{IEEEtranS}
\bibliography{root}
\end{document}